\pdfoutput=1

\documentclass[11pt]{article}

\usepackage[]{emnlp2021}

\usepackage{times}
\usepackage{latexsym}

\usepackage[T1]{fontenc}

\usepackage[utf8]{inputenc}

\usepackage{microtype}

\usepackage{microtype}
\usepackage{url}
\usepackage{graphicx}
\usepackage{float} 
\usepackage{subfigure}
\usepackage{amsmath}
\usepackage{amsfonts}
\usepackage{multirow}
\usepackage{colortbl}
\usepackage{arydshln}
\usepackage{array}
\usepackage{makecell}
\usepackage[utf8]{inputenc}
\usepackage{enumerate}

\usepackage{xcolor}
\usepackage{soul}
\newcolumntype{L}[1]{>{\raggedright\let\newline\\\arraybackslash\hspace{0pt}}m{#1}}
\setul{2pt}{2pt}
\usepackage{booktabs}

\usepackage{algorithm}
\usepackage{algpseudocode}
\usepackage{balance}

\usepackage{enumitem}
\setenumerate[1]{itemsep=0pt,partopsep=0pt,parsep=\parskip,topsep=5pt}
\setitemize[1]{itemsep=0pt,partopsep=0pt,parsep=\parskip,topsep=5pt}
\setdescription{itemsep=0pt,partopsep=0pt,parsep=\parskip,topsep=5pt}
\title{An Iterative Multi-Knowledge Transfer Network\\ for Aspect-Based Sentiment Analysis}

\author{
  Yunlong Liang\textsuperscript{1,2}\thanks{ \ \ Work was done when Yunlong Liang was interning at Pattern Recognition Center, WeChat AI, Tencent Inc, China.}, 
  Fandong Meng\textsuperscript{2}, 
  Jinchao Zhang\textsuperscript{2},
  \textbf{Yufeng Chen}\textsuperscript{1},\\ \textbf{Jinan Xu}\textsuperscript{1}\thanks{ \ \ Jinan Xu is the corresponding author.}
  \ and \textbf{Jie Zhou}\textsuperscript{2}\\
  \textsuperscript{1}Beijing Key Lab of Traffic Data Analysis and Mining, \\Beijing Jiaotong University, Beijing, China \\
  \textsuperscript{2}Pattern Recognition Center, WeChat AI, Tencent Inc, China \\
  \texttt{\{yunlongliang,chenyf,jaxu\}@bjtu.edu.cn} \\
  \texttt{\{fandongmeng,dayerzhang,withtomzhou\}@tencent.com} \\
}

\begin{document}
\maketitle
\begin{abstract}
Aspect-based sentiment analysis (ABSA) mainly involves three subtasks: aspect term extraction, opinion term extraction, and aspect-level sentiment classification, which are typically handled in a separate or joint manner. However, previous approaches do not well exploit the interactive relations among three subtasks and do not pertinently leverage the easily available document-level labeled domain/sentiment knowledge, which restricts their performances. To address these issues, we propose a novel \textbf{I}terative \textbf{M}ulti-\textbf{K}nowledge \textbf{T}ransfer \textbf{N}etwork (\textsc{IMKTN}) for end-to-end ABSA. For one thing, through the interactive correlations between the ABSA subtasks, our \textsc{IMKTN} transfers the task-specific knowledge from any two of the three subtasks to another one at the token level by utilizing a well-designed routing algorithm, that is, any two of the three subtasks will help the third one. For another, our \textsc{IMKTN} pertinently transfers the document-level knowledge, \emph{i.e.}, domain-specific and sentiment-related knowledge, to the aspect-level subtasks to further enhance the corresponding performance. Experimental results on three benchmark datasets demonstrate the effectiveness and superiority of our approach.
\end{abstract}

\section{Introduction}
\label{intro}
Aspect-based sentiment analysis (ABSA) has drawn increasing attention in the community, which includes three subtasks: \textbf{a}spect term \textbf{e}xtraction (AE), \textbf{o}pinion term \textbf{e}xtraction (OE) and aspect-level \textbf{s}entiment \textbf{c}lassification (SC). The first two subtasks aim to extract the aspect term and the opinion term appearing in one sentence, respectively. The goal of the SC subtask is to detect the sentiment polarity towards the extracted aspect term. 

Most existing studies generally handle each task separately~\cite{Tang:16a,Wang:16,hu-etal-2019-open} or take OE as auxiliary task for AE or SC~\cite{wang2017coupled,li2018aspect,he_acl2019}, where these separate approaches need to be pipelined or integrated
together for practical use. Recently, some researches point out that joint methods can achieve promising performance than separate ones, where only two subtasks are coupled, such as $\langle$AE, OE$\rangle$~\cite{wang2017coupled,dai-song-2019-neural} or $\langle$AE, SC$\rangle$~\cite{Luo2019doer,ijcai2019-762,he_acl2019,liang2020dependency}. More recently,~\citet{chen-qian-2020-relation} focus on modeling the interactive relations, \emph{i.e.}, bidirectional AE$\leftrightarrow$OE, unidirectional AE$\rightarrow$SC and unidirectional OE$\rightarrow$SC with a collaborative learning framework. To further enhance these subtasks, several researchers seek to the external accessible document-level corpora (containing domain-specific/sentiment-related knowledge\footnote{The two terms mean domain-relevant/sentiment-relevant linguistic knowledge, which are defined in~\cite{he_acl2019}}) due to the limited aspect-level data~\cite{dai-song-2019-neural,chen-qian-2019-transfer,He:18,he_acl2019}. As a better case, \citet{he_acl2019} merge the document-level domain-specific and sentiment-related knowledge together to enhance the AE and SC subtasks, where the two kinds of knowledge are indiscriminate.\footnote{We denote it a coarse way to use the domain-specific knowledge and sentiment-related knowledge together. By contrast, a fine-grained way is to separately and pertinently exploit the two kinds of knowledge to expert their advantages.}
Despite their effectiveness, we argue that the above methods are insufficient to yield satisfactory results for end-to-end ABSA task due to 1) they merely couple two subtasks or not modeling all bidirectional interactive relations among three subtasks (AE$\leftrightarrow$OE, AE$\leftrightarrow$SC and OE$\leftrightarrow$SC), and 2) the document-level domain-specific/sentiment-related knowledge is coarsely used, which is insufficient to exert their advantages.

\begin{figure}[t]
    \centering
    \includegraphics[width=0.48\textwidth]{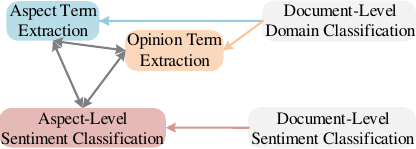}
    \caption{The interactive relations among three aspect-level subtasks (the left) and two document-level subtasks (the right), which are explicitly modeled through knowledge transferring. Three aspect-level subtasks are highly semantic correlated, and thus can incrementally facilitate one another through task-specific knowledge transfer. The domain-specific knowledge from domain classification is only transferred to aspect and opinion term extraction, and the sentiment-related knowledge from document-level sentiment classification is only for aspect-level sentiment classification.
    }
    \label{fig:case}
\end{figure}

First, the interactive relations among three aspect-level subtasks are mutually collaborative. For instance, in the sentence ``\emph{The \textbf{fish} is very \underline{delicious}.}'', the opinion term ``\emph{\underline{delicious}}'' indicates that the sentiment polarity of the aspect term ``\emph{\textbf{fish}}'' is {\em positive}, suggesting the strong interactive correlation among them. Conversely, given the aspect term ``\emph{\textbf{fish}}'' and its sentiment polarity {\em positive}, the word ``\emph{\underline{delicious}}'' rather than other words (\emph{e.g.}, ``\emph{very}'') in the sentence will be easily extracted as an opinion term. Therefore, the bidirectional relations between three aspect-level subtasks are closely related and they can incrementally promote one another, as shown in the left part of \autoref{fig:case}.

Second, the document-level corpora, which contain domain-specific and sentiment-related knowledge, should be pertinently utilized for enhancing the three aspect-level subtasks of ABSA. 
In fact, most aspect and opinion terms own distinct domain-specific properties~\cite{peng-etal-2018-cross} while sentiment polarities  (\emph{i.e.}, {\em positive}, {\em negative}, and {\em neutral}) are typically domain-invariant. For instance, the aspect term ``\emph{\textbf{fish}}'' and the opinion term ``\emph{\underline{delicious}}'' reflect distinct domain-specific characteristics, indicating that they belong to \emph{Restaurant} domain rather than \emph{Laptop} domain. Conversely, the domain-specific properties can help distinguish these aspect and opinion terms from other domains or background words (\emph{e.g.}, ``\emph{very}''). Therefore, the domain-specific knowledge should be pertinently leveraged to help identify the aspect term and the opinion term rather than on judging sentiment polarity. Meanwhile, the sentiment-related knowledge should be targeted at benefiting the SC subtask rather than the AE and OE subtasks, as shown in the right part of \autoref{fig:case}.

Therefore, we propose an \textbf{I}terative \textbf{M}ulti-\textbf{K}nowledge \textbf{T}ransfer \textbf{N}etwork ({\textsc{IMKTN}}) to fully exploit the interactive relations via transferring knowledge at both the token level and the document level for the ABSA task. Partially inspired by the superiority of capsule network in distinguishing different features by feature clustering~\cite{sabour2017dynamic}, we design a novel routing algorithm, which can mutually transfer task-specific knowledge among the three aspect-level subtasks, as illustrated in the left part of \autoref{fig:case}.  
Furthermore, \textsc{IMKTN} employs a more fine-grained way to pertinently transfer document-level knowledge to aspect-level subtasks, as shown in the right part of \autoref{fig:case}, where the knowledge from domain classification subtask only serves for the AE and OE subtasks while the knowledge from document-level sentiment classification subtask only helps the SC subtask. All multi-knowledge transfer processes are iteratively conducted for fully exploiting the knowledge in all tasks to enhance the ABSA task.

\label{sec:introduction}

In summary, our contributions are three-fold:
\begin{itemize}
\item We propose an iterative multi-knowledge transfer network for the ABSA task, which can well exploit the interactive relations via transferring the task-specific knowledge from any two of the three aspect-level subtasks to the third one for mutual promotion using a well-designed routing algorithm. 
\item We propose a more fine-grained way to pertinently transfer the document-level knowledge to further enhance the aspect-level tasks.
\item Our approach~\footnote{The code is publicly available at: \url{https://github.com/XL2248/IMKTN}} significantly outperforms the existing methods and achieves new state-of-the-art results on three benchmark datasets, namely SemEval14 (Restaurant14 and Laptop14)~\cite{Pontiki:14} and SemEval15 (Restaurant15)~\cite{Pontiki:15}.

\end{itemize}

\begin{figure*}[ht]
\centering
  \includegraphics[width = 0.9\textwidth]{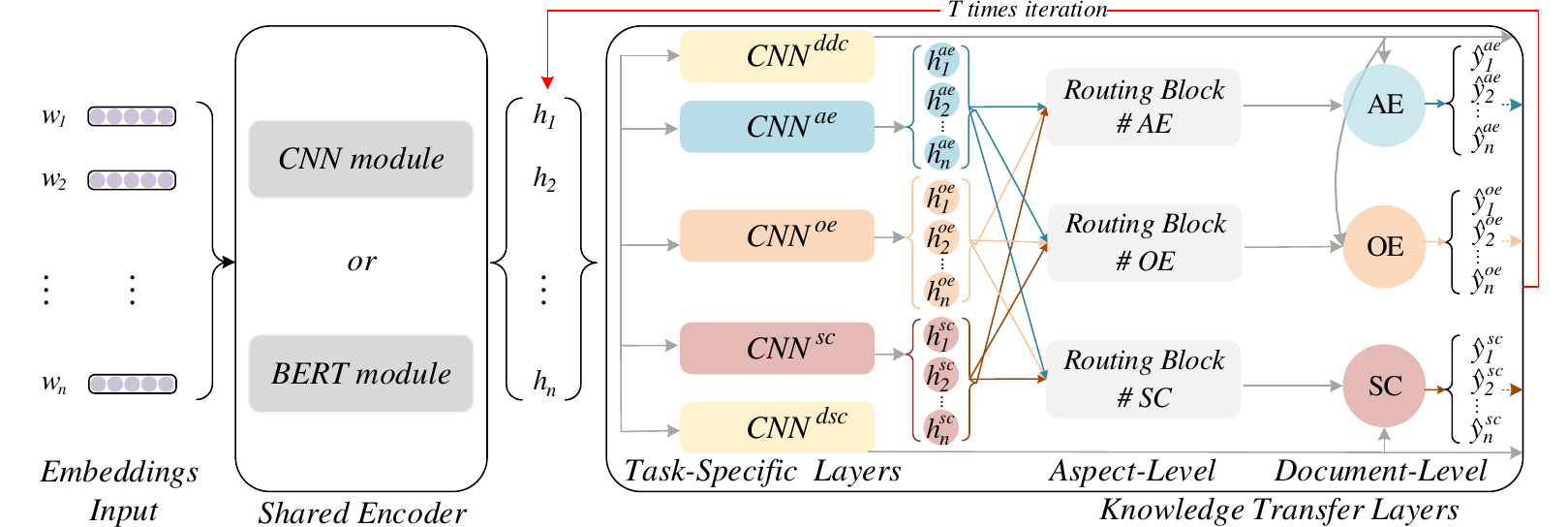}
\caption{The model architecture of \textsc{IMKTN}. AE: aspect term extraction; OE: opinion term extraction; SC: aspect-level sentiment classification. To fully exploit the inter-task correlations among the three aspect-level subtasks for mutual promotion, the knowledge from them is mutually transferred to each other via the ``\emph{Routing Block}''. Besides, the knowledge from {\em CNN$^{ddc}$} is only transferred to the AE and OE subtasks. The knowledge from {\em CNN$^{dsc}$} is only transferred to the SC subtask. In summary, all the multi-knowledge transfer processes are iteratively conducted for adequately exploiting the knowledge from all the subtasks to enhance the ABSA task.}
\label{fig:Architecture}
\end{figure*}

\section{Task Definition}
In this section, we formulate the aspect-level tasks and document-level tasks, where the document-level tasks are taken as auxiliary tasks for improving the aspect-level tasks.

\paragraph{Aspect-Level Tasks.} Following~\cite{chen-qian-2020-relation}, the ABSA task is formulated as three sequence labeling subtasks.
Given an input sentence $\mathrm{S}$=$\{w_i\}_{i=1}^n$ with $n$ words:
1) For the AE subtask, we aim to inference a tag sequence $\mathrm{ Y}^{ae}=\{{y}_i^{ae}\}_{i=1}^n$, where $y^{{ae}}_i \in \mathcal{Y}^{{ae}} =\{\texttt{BA},\texttt{IA},\texttt{O}\}$ denotes the \textbf{b}eginning and the \textbf{i}nside of an aspect term, and \textbf{o}ther words. 
2) For the OE subtask, we aim to inference a tag sequence $\mathrm{ Y}^{oe}=\{{y}_i^{oe}\}_{i=1}^n$, where $y^{{oe}}_i \in \mathcal{Y}^{{oe}}=\{\texttt{BP}, \texttt{IP}, \texttt{O}\}$ denotes the \textbf{b}eginning and the \textbf{i}nside of an opinion term, and \textbf{o}ther words.
3) For the SC subtask, we aim to inference a tag sequence $\mathrm{ Y}^{sc}=\{{y}_i^{sc}\}_{i=1}^n$, where $y^{{sc}}_i \in \mathcal{Y}^{{sc}}=\{\texttt{pos},\texttt{neg},\texttt{neu}\}$ denotes {\em positive}, {\em negative} and {\em neutral} sentiment polarities.

\paragraph{Document-Level Tasks.} This work contains two \textbf{d}ocument-level subtasks: \textbf{d}omain \textbf{c}lassification (DDC) and \textbf{s}entiment \textbf{c}lassification (DSC). For an input document $\mathrm{ D}=\{\mathrm{S}_1,\mathrm{S}_2,\dots,\mathrm{S}_m\}$ with $m$ sentences, the DDC and DSC aim to predict a domain label $\mathrm{ Y}^{ddc}\in\{Laptop, Restaurant\}$ and a sentiment label $\mathrm{Y}^{dsc}\in\mathcal{Y}^{{sc}}$, respectively. 

\section{Model}
As shown in \autoref{fig:Architecture}, the \textsc{IMKTN} consists of four parts: 1) Shared Encoder, for extracting n-gram features; 2) Task-Specific Layers, for capturing sentence representations; 3) Aspect-Level Knowledge Transfer, including three Routing Blocks, for fully transferring knowledge among the aspect-level subtasks for mutual reinforcing; and 4) Document-Level Knowledge Transfer, for pertinently transferring document-level knowledge to corresponding aspect-level tasks. Finally, multi-source information is aggregated for the next iteration.

\subsection{Shared Encoder}
We apply two modules to extract sentence features, 1) we adopt Convolutional Neural Network ({\em CNN})~\cite{Kim:14} as the feature extractor~\cite{kalchbrenner-etal-2014-convolutional}; 2) we investigate a more powerful encoder (\emph{i.e.}, BERT~\cite{bert}) as the backbone. The encoder is shared by the three aspect-level tasks and the two document-level tasks for providing common features. 

\subsection{Task-Specific Layers} 
Based on the Shared Encoder, 1) we design three aspect-level task-specific layers: {\em CNN}$^{ae}$, {\em CNN}$^{oe}$ and {\em CNN}$^{sc}$, aiming to generate aspect-related knowledge, opinion-related knowledge, and sentiment-related knowledge, respectively; and 2) two document-level task-specific layers: {\em CNN}$^{ddc}$ and {\em CNN}$^{dsc}$, for producing domain-specific features and sentiment features, respectively.

\subsection{Aspect-Level Knowledge Transfer}
As shown in \autoref{fig:Architecture}, we design an aspect-level knowledge transfer layer, consisting of three Routing Blocks, to take full advantage of the inter-task knowledge among the three aspect-level subtasks.

\begin{figure}[t]
\centering
  \includegraphics[width = 0.48\textwidth]{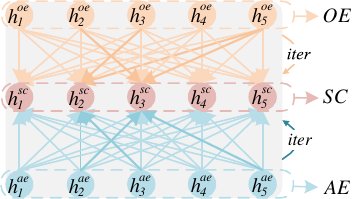}
\caption{An example of the internal structure of ``\emph{Routing Block \#SC}'' as shown in \autoref{fig:Architecture}. The knowledge of the AE and OE subtasks is transferred to the SC subtask through $iter$ rounds of iteration, that is, the AE and OE subtasks will help the SC subtask.}
\label{fig:routing}
\end{figure}

\paragraph{Routing Block.} The routing block serves for transferring knowledge among the aspect-level subtasks as shown in the ``\emph{Routing Block}'' part of \autoref{fig:Architecture}. Taking the ``\emph{Routing Block \#SC}'' for example, its internal structure is shown in \autoref{fig:routing}, in which the knowledge from AE and OE is transferred to SC for enhancing its performance via our routing algorithm. 
We use the same algorithm to transfer knowledge from OE and SC to AE through the ``\emph{Routing Block \#AE}'', from AE and SC to OE through the ``\emph{Routing Block \#OE}''.
In the conventional routing algorithm~\cite{sabour2017dynamic}, the high-level capsules are in a predefined fixed number, \emph{e.g.}, the total number of categories. While in our task, the high-level capsules are in dynamic numbers, where the number is determined by the sentence length. To this end, we propose a new routing algorithm, which is elaborated in detail below. 

We show the whole routing process in Algorithm~\autoref{algo} by taking ``transferring knowledge from OE to SC'' as example. Specifically, the inputs of Algorithm~\ref{algo} are the representation of OE (${\mathbf{h}^{oe}_i} \in \mathbb{R}^{d_h}$) and iteration number ($iter$) (line 1). The $b_{j|i}$ is the probability indicating that the representation of the $i$-th token in OE agrees to be routed to the representation of the $j$-th token in SC, which is initialized with zero (line 2). The $\mathbf{W}^p \in \mathbb{R}^{n \times d_h \times d_o}$ is position-aware transformation matrix, which is realized via adding positional encoding~\cite{attention17}, \emph{i.e.}, using $\mathrm{AddPos(\cdot)}$ function to obtain the shared transformation matrix $\mathbf{W}$ (line 3), where $\mathbf{W} \in \mathbb{R}^{d_h \times d_o}$. $PE_{(*)}$ is defined as: \begin{equation}\nonumber
\begin{split}
    PE_{(pos,2p)} = \mathrm{sin}(pos / 10000^{2p/d_{model}}), \\
    PE_{(pos,2p+1)} = \mathrm{cos}(pos / 10000^{2p/d_{model}}),
    \end{split}
\end{equation}
where $pos$ is token position in sentence, $p$ is the positional index of the dimension and $d_{model}$ is the input dimension. By doing so, the Algorithm can output capsules in dynamic numbers determined by the sentence length. The $\hat{\mathbf{u}}_{j|i}$ denotes the resulting opinion knowledge vector generated by multiplying the representation $\mathbf{h}^{oe}_i$ with the specially-designed transformation matrix $\mathbf{W}^p$ (line 4). 

During each iteration (line 5), the coupling coefficients between low-level capsules $\mathbf{h}^{oe}_i$ and high-level capsules $\mathbf{v}$ are obtained by applying the softmax function (line 6). Then $\mathbf{s}_j$ is calculated by aggregating all opinion vectors with $c_{j|i}$ as weights, voting for the sentiment polarity of the $j$-th token (line 7). After that the $\operatorname{squash}(\mathbf{s}_j) = \frac{||\mathbf{s}_j||^2}{1 + ||\mathbf{s}_j||^2} \frac{\mathbf{s}_j}{||\mathbf{s}_j||}$ scales the output $\mathbf{s}_j$ non-linearly to 0$\sim$1 (line 8). Once the $\mathbf{v}_j$ is updated in the current iteration, the probability $b_{j|i}$ becomes larger if the dot product $\hat{\mathbf{u}}_{j|i}\cdot\mathbf{v}^{oe}_{j}$ is large (line 9). That is, when the $\hat{\mathbf{u}}_{j|i}$ is more similar to the $\mathbf{v}^{oe}_j$, the dot product is larger, meaning that it is more likely to route this opinion knowledge to the $j$-th token and thus affects its sentiment polarity. Therefore, larger $b_{j|i}$ will lead to a larger agreement value $c_{j|i}$ between the opinion knowledge of the $i$-th token and the sentiment representation of the $j$-th token in the next iteration. In contrast, it generates low $c_{j|i}$ when there is no correlation between $\hat{\mathbf{u}}_{j|i}$ and $\mathbf{v}^{oe}_{j}$. 
After $iter$ rounds of iteration, agreement values learned via the routing process ensure the opinion knowledge will be sent to the appropriate sentiment representation.

Similarly, we obtain the knowledge $\mathbf{v}^{ae}_j$, which is transferred from AE to SC, indicating which token should be correctly labeled with the sentiment polarity. Then the knowledge from AE and OE subtasks is combined as follows:
\begin{equation}\nonumber
  \label{eq:knowledge-representation}
  \begin{split}
  \mathbf{h}^{sc}_j &= \mathrm{Concat}(\mathbf{h}^{sc}_j,\mathbf{v}^{ae}_j,\mathbf{v}^{oe}_j),
  \end{split}
\end{equation}
where $\mathbf{h}^{sc}_j \in \mathbb{R}^{d_h+2d_o}$ is the $j$-th hidden state of the SC subtask (we set dimension size of all output capsules to $d_o$). 
\begin{algorithm}[t]
\resizebox{7cm}{!}{
\caption{Routing}
\begin{minipage}{1.16\linewidth}
\begin{algorithmic}[1]
\Procedure{Routing algorithm}{${\mathbf{h}^{oe}_i}$, $iter$}
    \State {$\forall i \in OE, \forall j \in SC, 1 \leq i, j \leq n, {b}_{j|i}\leftarrow0$.}
    \State {$\mathbf{W}^p = \operatorname{AddPos}(\operatorname{tile}\footnote{The tile operation of Tensorflow~\cite{DBLP:journals/corr/AbadiBCCDDDGIIK16}.}(\mathbf{W}, n), PE_{(pos,2p)}, PE_{(pos,2p+1)}$})
    \State {$\hat{\mathbf{u}}_{j|i} = \mathbf{h}^{oe}_i\mathbf{W}_{ij}^p$} 
    \For {$iter$~iterations}
        \State {$\forall i \in OE$: $\mathbf{c}_i\leftarrow\operatorname{softmax}(\mathbf{b}_{i})$}
        
        \State {$\forall j \in SC$: $\mathbf{s}_j \leftarrow \Sigma_i c_{j|i}\hat{\mathbf{u}}_{j|i}$}
        
        \State {$\forall j \in SC: \mathbf{v}^{oe}_j \leftarrow \operatorname{squash}(\mathbf{s}_j)$}
        
        \State {$\forall i \in OE. \forall j \in SC: {b}_{j|i} \leftarrow {b}_{j|i} + \hat{\mathbf{u}}_{j|i}\cdot\mathbf{v}^{oe}_{j}$}
        
	    \EndFor
	\State Return $\mathbf{v}^{oe}_j$
\EndProcedure
\end{algorithmic}\label{algo}\label{Al_routing}
\end{minipage}
}
\end{algorithm}

Through the process above, the multi-knowledge transfer in ``\emph{Routing Block \#SC}'' is finished, which determines the sentiment polarity of each token in SC. Similarly, we achieve multi-knowledge transfer in ``\emph{Routing Block \#OE}'' and ``\emph{Routing Block \#AE}'' in \autoref{fig:Architecture}. By doing so, three aspect-level subtasks are interacted with one another to fully exploit the inter-task correlations. 

\subsection{Document-Level Knowledge Transfer}
We design the following two ways to pertinently transfer document-level knowledge to corresponding aspect-level tasks. (1) We transfer domain-specific knowledge ($a_i^{ddc(t)}$ and $a_i^{ddc(t)}$) from the DDC subtask to the AE and OE subtasks: 
\begin{equation}\nonumber
\resizebox{1.0\hsize}{!}{$
\begin{split}
    \mathbf{h}^{q(t+1)}_i= 
    &f_1([\mathbf{h}^{q(t)}_i; \hat{\mathbf{y}}_i^{ae(t)}; \hat{\mathbf{y}}_i^{oe(t)};\hat{\mathbf{y}}_i^{sc(t)}; y_i^{ddc(t)};  a_i^{ddc(t)}]),
\end{split}
$}\label{update_hidden2}
\end{equation}
where $q \in \{{ae}, oe\}$, $t$ is the iteration number ($0 \le t\le T$), $[\cdot;\cdot]$ denotes concatenation operation, $f_1(\cdot)$ is fully-connected layer and $\mathbf{\hat{y}}_i^{o(t)}$ is the prediction on the $i$-th token at the $t$-th iteration, which is proved helpful in~\cite{he_acl2019}, $o \in \{ae, oe, sc\}$. 
(2) We transfer sentiment-related knowledge ($\hat{\mathbf{y}}^{dsc(t)}$ and $a^{dsc(t)}_i$) from the DSC subtask to the SC subtask:
\begin{equation}\nonumber
\resizebox{0.9\hsize}{!}{$
\begin{split}
    \mathbf{h}^{sc(t+1)}_i= 
    &f_2([\mathbf{h}_i^{sc(t)}; \hat{\mathbf{y}}^{ae(t)}_i; \hat{\mathbf{y}}^{oe(t)}_i;\hat{\mathbf{y}}^{sc(t)}_i;\\ &\hat{\mathbf{y}}^{dsc(t)}; a^{dsc(t)}_i]),
\end{split}
$}\label{update_hidden1}
\end{equation}
where $f_2(\cdot)$ is fully-connected layer. $a_i^{s(t)}$ ($s \in \{ddc, dsc \}$) is the self-attention weight (at the document level): 
\begin{equation}\nonumber
    a_i^{s(t)} = \frac{\mathrm{exp}(\mathbf{h}_i^{s(t)}\mathbf{W}^{s})}{\sum_{k=1}^{n} \mathrm{exp}(\mathbf{h}_k^{s(t)}\mathbf{W}^{s})},
\end{equation}
where $\mathbf{W}^{s}$ is the trainable parameter. The document representation is computed by 
\begin{equation}\nonumber
\mathbf{h}^{s(t)} = {\sum_{i=1}^{n}{a}_i^{s(t)}\mathbf{h}^{s(t)}_i}.
\end{equation}
Then a fully-connected layer with softmax function is applied to map $\mathbf{h}^{s(t)}$ to $\mathbf{\hat{y}}^{s(t)}$. 

Overall, the \textsc{IMKTN} can fully perform knowledge transfer via the routing algorithm and pertinently incorporate the document-level knowledge to enhance the corresponding aspect-level tasks through such $T$ rounds of iteration. 

\subsection{Training}
For training, we minimize the loss on each token of aspect-level tasks and each instance of document-level tasks with the cross-entropy function. The aspect-level loss functions are written as follows:
\begin{equation}\nonumber
\resizebox{.7\hsize}{!}{$
    \begin{split}
     \mathcal{J}_a = & \lambda_1\mathcal L_{ae} + \lambda_2\mathcal L_{oe} + \lambda_3\mathcal L_{sc}, \\
     \mathcal L_{o} = &\frac{1}{n} \sum\limits_{i=1}^{n} (min (- \sum_{r=0}^{C_1}\mathbf{y}_{i,r}^{o}\mathrm{log}(\hat{\mathbf{y}}_{i,r}^{o(T)})),
    \end{split}
$}
\label{aspect_loss}
\end{equation}
where $\lambda_1,\lambda_2$ and $\lambda_3$ are discount coefficients, $o \in \{ae, oe, sc \}$, $n$ is the sentence length, $C_1$ is the class number, $\mathbf{y}_{i,r}^{o}$ denotes the ground-truth and $\hat{\mathbf{y}}_{i,r}^{o(T)}$ denotes the predictions with $T$ times iteration. The document-level loss functions are formulated as follows:
\begin{equation}\nonumber
\resizebox{.6\hsize}{!}{$
\begin{split}
  \mathcal{J}_d &= \lambda_4\mathcal L_{ddc} + \lambda_5\mathcal L_{dsc}, \\
  \mathcal L_{s} &=  min (- \sum_{r=0}^{C_2}\mathbf{y}_{r}^{s} \mathrm{log}(\hat{\mathbf{y}}_{r}^{s(T)})),
\end{split}
$}
\end{equation}
where $\lambda_4$ and $\lambda_5$ are discount coefficients, $s \in \{ddc, dsc\}$, $C_2$ is the class number, $\mathbf{y}_{r}^{s}$ denotes the ground-truth and $\hat{\mathbf{y}}_{r}^{s(T)}$ denotes the predictions after $T$ times iteration.

For training the whole model, we firstly train the network with document-level tasks for a few epochs to generate reasonable features for aspect-level tasks. Then we train the network on the aspect-level and document-level corpus alternately, to minimize the corresponding loss.

\renewcommand{\arraystretch}{1.1}
\begin{table}[t]
\centering
\small
\scalebox{0.87}{
\setlength{\tabcolsep}{0.75mm}{
\begin{tabular}{ll|ccc|ccc}
\toprule 
&\multirow{2}{*}{Datasets} & \multicolumn{3}{c}{Train} &  \multicolumn{3}{|c}{Test}\\\cline{3-8}
&& \#sent & \#aspect & \#opinion & \#sent & \#aspect & \#opinion\\\hline
D1&Restaurant14 &3,044 &3,699 &3,484 &800 &1,134 &1,008\\
D2&Laptop14 &3,048 &2,373 &2,504 &800 &654 &674 \\
D3&Restaurant15 &1,315 &1,199 &1,210 &685 &542 &510\\
\bottomrule
\end{tabular}}}
\caption{Dataset statistics. \#sent: sentences, \#aspect: aspect terms and \#opinion: opinion terms.}\label{datasets}
\end{table}
\section{Experiments}
\subsection{Experimental Settings}
\paragraph{Datasets.} We evaluate our model on three benchmark datasets from SemEval 2014 (Restaurant14 and Laptop14)~\cite{Pontiki:14} and SemEval 2015 (Resaurant15)~\cite{Pontiki:15}, the data statistics of which is shown in \autoref{datasets}. The opinion terms of these three datasets are annotated by~\citeauthor{rncrf}~\shortcite{rncrf}. We adopt two document-level datasets from~\citeauthor{he_acl2019}~\shortcite{he_acl2019}, which include 30k instances of Yelp restaurant domain and 30k instances of Amazon electronic domain, respectively. We merge the two datasets with domain labels for domain classification. We use the Yelp data when training on D1 and D3, and use the Amazon data for D2, due to the domain-specific properties.


\begin{table*}[t]
\centering
\newcommand{\tabincell}[2]{\begin{tabular}{@{}#1@{}}#2\end{tabular}}
\small
\setlength{\tabcolsep}{0.800mm}{
\begin{tabular}{ll|cccc|cccc|cccc}
\toprule
&\multirow{ 2}{*}{Models} & \multicolumn{4}{c|}{Restaurant14 (D1)} &  \multicolumn{4}{c|}{Laptop14 (D2)} & \multicolumn{4}{c}{Restaurant15 (D3)}\\\cline{3-14}
& & F1-ae & F1-oe & F1-sc & F1-absa & F1-ae & F1-oe & F1-sc & F1-absa & F1-ae & F1-oe & F1-sc & F1-absa\\\hline
M1&CMLA-TNet$^*$ &81.91 &83.84 &69.69  &64.49     &77.49 &76.06 &68.30 &55.94    &67.73 &70.56 &62.27 &55.00\\
M2&CMLA-TCap$^*$ &82.45 &82.67 &72.23  &65.34     &76.80 &77.33 &69.52  &55.56    &68.55 &71.07 &66.45 &55.47\\
M3&DECNN-TNet$^*$ &82.79 &- &70.45  &65.80     &79.38 &- &68.69 &57.39     &68.52 &- &62.41 &55.69\\
M4&DECNN-TCap$^*$ &82.79 &- &71.77 &66.84      &79.38 &- &69.61 &57.71    &68.52 &- &63.60 &56.22\\\cdashline{1-14}[4pt/2pt]
M5&MNN$^*$  &83.05 &84.55 &68.45  &63.87       &76.94 &77.77 &65.98 &53.80     &70.24 &69.38 &57.90  &56.57\\
M6&INABSA$^*$ &83.92 &84.97 &68.38  &66.60      &77.34 &76.62 &68.24 &55.88    &69.40 &71.43 &58.81 &57.38\\\cdashline{1-14}[4.0pt/2pt]
M7&DOER$^*$ &84.63 &- &64.50  &68.55      &80.21 &- &60.18 &56.71    &67.47 &- &36.76 &50.31\\
M8&Span-based&84.13 &- &69.73  &68.22   &78.43 &- &69.77 &57.57    &69.96 &- &59.95 &58.97\\
M9&IMN$^{\natural}$  &{83.33} &{85.61} &{75.66}  &{69.54}    &77.96 &77.51 &72.02 &{58.37}   &{70.04} &71.94 &{71.76} &{59.18}\\
M10&DREGCN$^{\natural}$ &{85.93} &{86.05} &{73.32} &{70.21}    &79.45 &75.40 &73.46  &{61.60}      &{71.00} &70.55  &{73.35} &{61.06}\\
M11&RACL$^*$ &{85.37} &{85.32} &{74.46}  &{70.67}   &81.99 &79.76 &71.09 &{60.63}  &{72.82} &\bf{78.06} &{68.69} &{60.31}\\ 
M12& \textsc{IMKTN}-GloVe &\bf{87.91}$^{\dagger}$ &\bf{87.65}$^{\dagger}$  &\bf{76.66}$^{\dagger}$  &\bf{72.80}$^{\dagger}$    &\bf{83.19}$^{\dagger}$ &\bf{81.82}$^{\dagger}$ &\bf{74.93}$^{\dagger}$ &\bf{62.96}$^{\dagger}$    &\bf{74.96}$^{\dagger}$ &{74.48} &\bf{75.39}$^{\dagger}$ &\bf{63.17}$^{\dagger}$ \\\hline
M13&{SPAN-BERT}$^{*}$ &{86.71} &- &{71.75} &{73.68}    &82.34 &- &62.50  &{61.25}      &{74.63} &-  &{50.28} &{62.29}\\
M14&{IMN-BERT}$^{*}$ &{84.06} &85.10 &{75.67} &{70.72}    &77.55 &81.00 &75.56  &{61.73}      &{69.90} &73.29  &{70.10} &{60.22}\\
M15&{DREGCN-BERT}$^{\natural}$ &{87.00} &86.95 &75.79 &{72.60}    &79.78 &79.21 &76.37  &{63.04}      &{73.30} &72.60  &{73.02} &{62.37}\\
M16&{RACL-BERT}$^{*}$ &{86.38} &{87.18} &\bf{81.61} &{75.42}    &81.79 &79.72 &73.91  &{63.40}      &{73.99} &76.00  &{74.91} &{66.05}\\
M17&\textsc{IMKTN}-BERT &\bf{87.13}$^{\dagger}$ &\bf{88.62}$^{\dagger}$ &{81.35}&\bf{76.75}$^{\dagger}$    &\bf{83.89}$^{\dagger}$ &\bf{81.90}$^{\dagger}$ &\bf{76.42}$^{\dagger}$ &\bf{65.74}$^{\dagger}$   &\bf{74.63} &\bf{76.79}$^{\dagger}$  &\bf{76.85}$^{\dagger}$&\bf{68.33}$^{\dagger}$\\
\bottomrule
\end{tabular}}
\caption{Model comparison. We separate the results into the GloVe-based (M1$\sim$M12) and BERT-based (M13$\sim$M17) methods for fair comparison. Following RACL~\cite{chen-qian-2020-relation}, we report average results over 5 runs with random initialization. The results with the symbol ``$^*$'' refer to RACL. ``$^{\natural}$'' indicates that the results are referred to the original paper. ``$^{\dagger}$'' denotes our method is statistically significant~\cite{koehn-2004-statistical} better than RACL ({\em p}-value \textless \ 0.05), which is the best previous model.  
}\label{main results} 
\end{table*}

\paragraph{Implementation Details.} For fair comparison, we train our models with the same settings as comparison models~\cite{chen-qian-2020-relation}. We tune the iteration number $T$ and the routing number $iter$ on each validation set. More implementation and tuning details are given in Appendix A and B. 

\paragraph{Evaluation Metrics.} Following~\cite{chen-qian-2020-relation}, four metrics are applied for evaluation, and the average score over 5 runs with random initialization is reported in all experiments. We use \textbf{F1-ae}, \textbf{F1-oe} and \textbf{F1-sc} to denote the F1-score of each subtask. We use F1-score denoted as \textbf{F1-absa} to measure the complete ABSA,\footnote{Following~\cite{chen-qian-2020-relation}, we use the predicted sentiment of the first word as the SC result if an aspect term has multiple words. Besides, aspect terms with \emph{conflict} sentiment labels are ignored. All baseline models apply the same setting for fair comparison.} where an extracted aspect term is taken as correct only when the span and the sentiment are both correct. 

\subsection{Comparison Models}
To validate the performance of our proposed model on the ABSA task, we conduct contrast experiments with the following methods:

    \noindent\textbf{Pipeline Models.} We respectively select two top performing models for AE: CMLA~\cite{wang2017coupled} and DECNN~\cite{xu_acl2018}, and SC: TNet~\cite{li2018transformation} and TCap~\cite{chen-qian-2019-transfer}, to construct 2 × 2 pipeline baselines. {SPAN-BERT}~\cite{hu-etal-2019-open} utilizes $BERT_{LARGE}$ as backbone networks for AE and SC subtasks.
    
    \noindent\textbf{Integrated Models.} {MNN}~\cite{Wang18} and {INABSA}~\cite{li2019unified}: Both models handle the aspect term-polarity co-extraction as a sequence labeling problem with a unified tagging scheme.
    
    \noindent\textbf{Joint Models.} The joint models including {DOER}~\cite{Luo2019doer}, {Span-based}~\cite{ijcai2019-762}, {IMN}~\cite{he_acl2019},  {DREGCN}~\cite{liang2020dependency}, and {RACL}~\cite{chen-qian-2020-relation} are used to compare with ours, which are introduced in~\autoref{intro} part.

For fair comparison, we validate \textsc{IMKTN} based on two encoders. 1) Based on CNN, we use GloVe embeddings~\cite{glove:14} and denote it as \textsc{IMKTN}-GloVe. 2) Based on $BERT_{LARGE}$~\cite{bert}, we fine-tune it for ABSA, denoted as \textsc{IMKTN}-BERT).

\subsection{Main Results}
Results in \autoref{main results} are divided into four groups: M1$\sim$M4, M5$\sim$M6, and M7$\sim$M12 are GloVe-based pipeline, integrated, and joint models, respectively. M13$\sim$M17 are BERT-based models.

1) Among all GloVe-based models (M1$\sim$M12), our \textsc{IMKTN}-GloVe significantly surpasses other baselines in most cases, and achieves 2.13\%, 2.33\%, and 2.86\% absolute gains over RACL in terms of the overall metric F1-absa on three datasets. This suggests that the inter-task correlations and document-level knowledge have an overall positive impact on these subtasks, and demonstrates the superiority of our model. Furthermore, \textsc{IMKTN}-GloVe also obtains the best or the second best results on all subtasks, which further shows the effectiveness of our model. Another observation is that the joint models (M7$\sim$M12) perform better than pipeline and integrate models (M1$\sim$M6).

2) All BERT-based models get higher results than GloVe-based models thanks to the large-scale external knowledge (M13$\sim$M17 vs. M1$\sim$M12). Among all BERT-based Models, we observe that \textsc{IMKTN}-BERT significantly outperforms other BERT-based models, which suggests the effectiveness of our approach by transferring multi-source knowledge even based on the strong baseline and yields new state-of-the-art results on most metrics.

\begin{table}[t]
\centering
\newcommand{\tabincell}[2]{\begin{tabular}{@{}#1@{}}#2\end{tabular}}
\small
\scalebox{0.84}{
\setlength{\tabcolsep}{1.5mm}{
\begin{tabular}{c|l|c|c|c}
\toprule
&\multirow{1}{*}{Models} & \multicolumn{1}{c|}{D1} &  \multicolumn{1}{c|}{ D2} & \multicolumn{1}{c}{D3}\\\cline{1-5}
\multirow{4}{*}{\tabincell{c}{Aspect-Opinion Pair}}
&IMN$^{\natural}$ &54.94 &54.87 &56.45\\
&DREGCN$^{\natural}$ &53.76 &54.89 &55.23 \\
&RACL$^{\natural}$ &54.67 &54.75 &56.74 \\
&\textsc{IMKTN}-D &\bf{56.74}$^{\dagger}$ &\bf{56.60}$^{\dagger}$ &\bf{58.32}$^{\dagger}$ \\\hline
\multirow{4}{*}{\tabincell{c}{Aspect-Opinion-Sentiment\\ Triplet}}
&IMN$^{\natural}$ &50.95 &41.21 &45.65 \\
&DREGCN$^{\natural}$ &49.32 &41.97 &44.38 \\
&RACL$^{\natural}$ &50.65 &41.55 &45.45\\
&\textsc{IMKTN}-D &\bf{52.45}$^{\dagger}$ &\bf{44.82}$^{\dagger}$ &\bf{48.50}$^{\dagger}$ \\
\bottomrule
\end{tabular}}}
\caption{F1 scores (\%). The aspect-sentiment pair results are shown in \autoref{main results}, \emph{i.e.}, F1-absa score. ``$^{\natural}$'': results are generated by running their official code. ``$^{\dagger}$'': significantly better than RACL ({\em p}-value \textless \ 0.05).}\label{main_Statistical_Supports} 
\end{table}
\begin{table}[t]
\centering
\newcommand{\tabincell}[2]{\begin{tabular}{@{}#1@{}}#2\end{tabular}}
\small
\scalebox{0.9}{
\setlength{\tabcolsep}{2.5mm}{
\begin{tabular}{l|l|c|c|c}
\toprule
\#&Methods & F1-ae & F1-oe & F1-sc \\\hline
0&Coarse way &81.06 &85.02 &{65.44}  \\
1&Fine-Grained way &\bf{82.25} &\bf{86.36} &\bf{68.80}  \\
\bottomrule
\end{tabular}}}
\caption{F1 (\%) on the validation set of D1.}\label{doc-exp} 
\end{table}

\section{Analysis and Discussion}
\subsection{Whether Three Aspect-Level Subtasks Promote Each Other?}
We evaluate the aspect-opinion pair F1 and aspect-opinion-sentiment triplet F1 on the test set~\cite{fan-etal-2019-target,peng2019knowing,xu-etal-2020-position},\footnote{Table 9 (in Appendix) shows an example for explaining what are aspect-sentiment pair, aspect-opinion pair, and aspect-opinion-sentiment triplet.} for verifying whether the multi-knowledge transferring can help each other. The results are shown in \autoref{main_Statistical_Supports}, where \textsc{IMKTN}-D denotes removing all document-level knowledge transferring. We can see that our \textsc{IMKTN}-D can surpass the comparison models by a large margin under two settings. Particularly, in the aspect-opinion-sentiment triplet setting, \textsc{IMKTN}-D significantly outperforms other baselines, suggesting that inter-task knowledge transferring has an overall positive impact on these aspect-level subtasks and hence the aspect-level subtasks indeed can promote each other.

\begin{table}[t]
\centering
\newcommand{\tabincell}[2]{\begin{tabular}{@{}#1@{}}#2\end{tabular}}
\small
\scalebox{0.9}{
\setlength{\tabcolsep}{2.0mm}{
\begin{tabular}{l|l|c|c|c}
\toprule
\#&Models & D1 & D2 & D3 \\\hline
0&\emph{w/o} AE KT &1.05/1.98$\downarrow$ &1.56/1.44$\downarrow$ &1.45/3.45$\downarrow$  \\
1&\emph{w/o} OE KT  &0.98/0.45$\downarrow$ &0.96/0.52$\downarrow$ &1.13/2.09$\downarrow$ \\
2&\emph{w/o} SC KT &1.89/2.78$\downarrow$ &1.83/2.23$\downarrow$ &2.38/4.54$\downarrow$\\
3&\emph{w/o} DDC &1.88/2.03$\downarrow$ &1.87/1.82$\downarrow$ &1.54/3.37$\downarrow$\\
4&\emph{w/o} DSC &2.37/2.77$\downarrow$ &2.13/2.43$\downarrow$ &2.87/5.04$\downarrow$\\
\bottomrule
\end{tabular}}}
\caption{Ablation study. ``$\downarrow$'' denotes a performance drop of ``\textsc{IMKTN}-GloVe/\textsc{IMKTN}-BERT'' on the validation set (F1-absa). ``KT'': knowledge transferring.
}\label{ablation}
\end{table}

\subsection{Whether Pertinently Transferring Document-Level Knowledge Helps Aspect-Level Subtasks More?} 
In \autoref{doc-exp}, the ``Coarse way''~\cite{he_acl2019} indicates that the knowledge from DDC and DSC is merged to indistinguishably enhance all aspect-level tasks. By contrast, the ``Fine-Grained way'' is to pertinently transfer the knowledge, \emph{\emph{i.e.}}, the knowledge from DDC only transferred to AE and OE subtasks, and the knowledge from DSC only transferred to SC subtask. The results show that pertinently transferring document-level knowledge helps aspect-level subtasks more, which is consistent with our intuition that the domain-specific knowledge prefers to promote the AE and OE subtasks, and the sentiment-related knowledge tends to improve the SC subtask. Therefore, a fine-grained way is very necessary to enhance the ABSA. 

\subsection{Ablation Study} 
\label{ablation_study}
\autoref{ablation} shows the impact of different knowledge, where we remove one knowledge at a time. We conclude that: (1) once any of the aspect-level subtask knowledge transfer is removed (rows 0$\sim$2), scores on three benchmark datasets decrease under the both setting (\emph{i.e.}, GloVe and $BERT_{LARGE}$), showing that the three aspect-level subtasks are highly semantically correlated and thus can incrementally boost one another. (2) we also observe obvious drops when removing the document-level knowledge, especially when the DSC subtask is removed, suggesting that pertinently transferring the document-level knowledge significantly benefits the corresponding aspect-level tasks (rows 3$\sim$4). 

\subsection{Why using Capsule Network?}
In our preliminary experiments, we conduct some experiments to investigate how to effectively transfer knowledge between different tasks. The results are shown in \autoref{smallexperiment}, where the capsule network (row 3) performs the best. The reason is capsules in adjacent layers connected by dynamic routing, which has the ability to distinguish different features by feature clustering~\cite{sabour2017dynamic}. This coincides with our motivation, \emph{i.e.}, 
transferring related features from two subtasks to the third one through the bidirectional interactive relations for mutual promotion (feature clustering). However, other methods (rows 0$\sim$2) have no such dynamic routing mechanism and thus cannot dynamically conduct feature extraction and clustering, leading to unsatisfactory results.  Therefore, we select the capsule network.

\subsection{Case Study and Visualization}
\begin{figure}[t]
    \centering
    \includegraphics[width=0.46\textwidth]{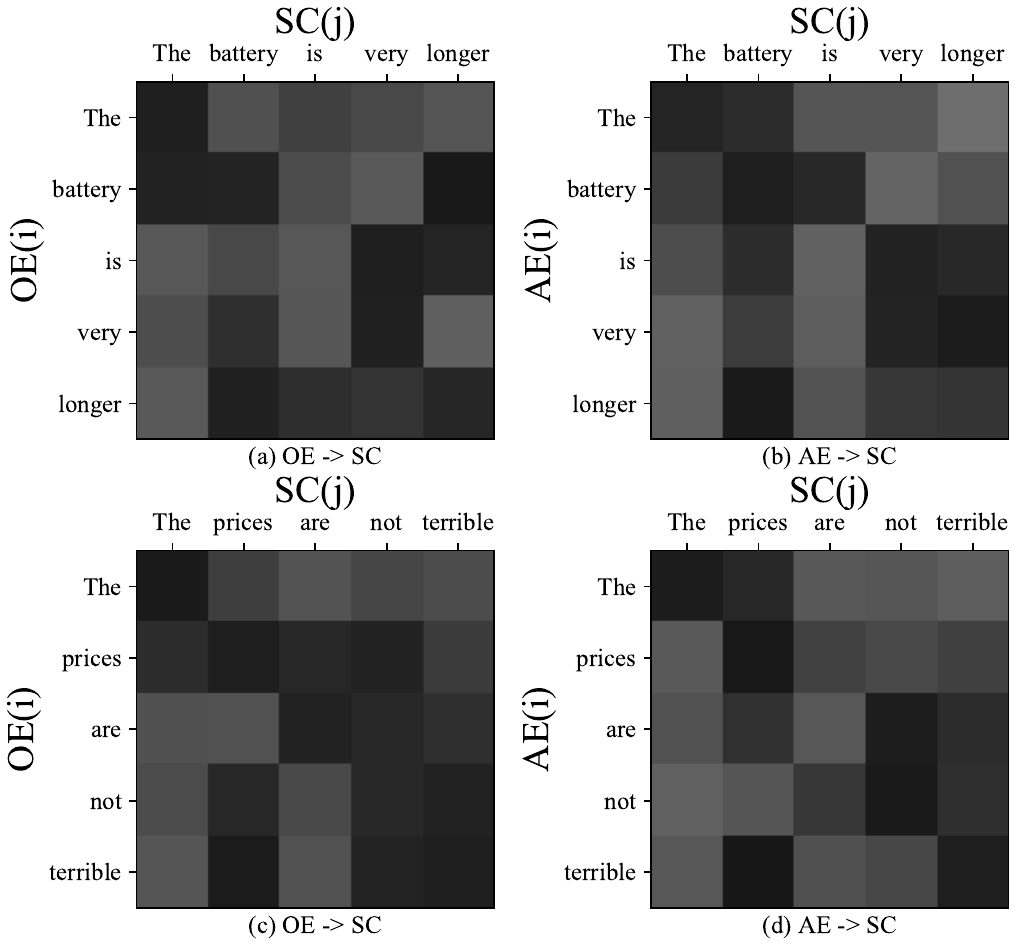}
    \caption{Visualization of $c_{j|i}$. The darker the color is, the more knowledge is transferred. 
    }
    \label{fig:v2}
\end{figure}

To provide an understanding of how the multi-knowledge transfer works, in \autoref{fig:v2},\footnote{Both two examples are taken from the Laptop 14 dataset.} we take the knowledge transfer from OE and AE to SC for example to visualize the agreement value $c_{j|i}$. \autoref{fig:v2}(a) and~\autoref{fig:v2}(c) are the cases of transferring knowledge from OE to SC. \autoref{fig:v2}(a) shows that the knowledge of opinion term ``\emph{longer}'' from the OE subtask is mainly sent to aspect term ``\emph{battery}'' of the SC subtask and \autoref{fig:v2}(c) shows the same phenomenon (the knowledge of opinion term ``\emph{not terrible}'' from the OE subtask is mainly sent to the aspect term ``\emph{prices}'') though it is a negation sentence, indicating that the opinion word affects the sentiment polarity of the aspect term, \emph{i.e.}, the former (AE) is naturally correlated with the latter (SC). Particularly, in \autoref{fig:v2}(c), negation information can be effectively transferred to the aspect term ``\emph{prices}'' via the routing algorithm and affects its sentiment polarity. \autoref{fig:v2}(b) and~\autoref{fig:v2}(d) are the cases of transferring knowledge from AE to SC, showing that the aspect-related knowledge is mainly transferred to the aspect term ``\emph{battery}'' and ``\emph{prices}'', voting for them to be aspect terms. Therefore, the AE subtask can help the aspect-level sentiment classification to judge whether the word should own sentiment polarity or not. Besides, we also present thorough error analysis in Appendix C.

\begin{table}[t]
\centering
\small
\scalebox{0.9}{
\setlength{\tabcolsep}{2.9mm}{
\begin{tabular}{c|l|c|c|c}
\toprule 
\#&Methods &D1 &D2 &D3\\\hline
0&Concat &60.56  &50.11&67.73 \\
1&LSTM &60.77  &51.19&66.93 \\
2&Attention &{61.36} &52.49&68.02  \\ 
3&Capsule &\bf{62.89} &\bf{54.10}&\bf{70.36}  \\
\bottomrule
\end{tabular}}}
\caption{F1-absa (\%) on the validation set. Apart from using ``\emph{Routing Blocks}'' in \autoref{fig:Architecture}, we also try the following three methods. \romannumeral1): We directly concatenate the task-specific features (row 0). \romannumeral2) We use an LSTM to sequentially read the task-specific features for transferring knowledge (row 1). \romannumeral3) We apply attention to calculate the score between the task-specific features, and then take the score as the weight to conduct the task-specific knowledge transferring (row 2).
}\label{smallexperiment}
\end{table}

\section{Related Work}
\paragraph{Aspect-Based Sentiment Analysis.} %
Existing models typically handle the ABSA task independently or jointly. Apparently, separately treating each subtask cannot exploit the inter-task correlations, leading to restricted performances, such as AE (\citealp{qiu2011opinion,liu-etal-2013-syntactic,liu-etal-2014-extracting,liu-etal-2015-fine,wdbem,li2017deep,li2018aspect,angelidis-lapata-2018-summarizing,ma-etal-2019-exploring}, etc) and SC (\citealp{dong-etal-2014-adaptive,nguyen-shirai-2015-phrasernn,vo2015target,Chen:17,WangS:18,ma2018targeted,hu-etal-2019-constrained,liang-etal-2019-novel,bao-etal-2019-attention,sun-etal-2019-aspect,tang-etal-2019-progressive,tang-etal-2019-progressive,xu2020dombert}, etc). By contrast, the integrated or joint methods~\citep{rncrf,mitchell-etal-2013-open,zhang-etal-2015-neural,li2017learning,schmitt-etal-2018-joint,li-etal-2019-exploiting,lin-yang-2020-shared,liang2020dependency,chen-qian-2020-relation} can model the interactive correlations and thus achieve promising results. 
Different from above studies, we focus on exploiting the inter-task correlations among the three aspect-level subtasks and thus incrementally boost one another. Besides, we observe the task characteristics and then use the document-level corpora to pertinently help the corresponding aspect-level subtasks.

\paragraph{Capsule Network.} Capsule network~\cite{sabour2017dynamic} has been widely applied in many natural language processing tasks. In ABSA,~\citeauthor{Wang:2019:ASA:3308558.3313750}~\shortcite{Wang:2019:ASA:3308558.3313750} focus on building multiple capsules for aspect category sentiment analysis, which do not employ the routing procedure.  ~\citeauthor{chen-qian-2019-transfer}~\shortcite{chen-qian-2019-transfer} construct a transfer capsule network for transferring semantic knowledge from DSC to SC via sharing the encoder, which utilizes the vanilla capsule network only for the SC subtask.  ~\citeauthor{du-etal-2019-capsule}~\shortcite{du-etal-2019-capsule} combine capsule network with interactive attention to model the interactive relationship between the given aspect term and context for the SC subtask. ~\citeauthor{jiang-etal-2019-challenge}~\shortcite{jiang-etal-2019-challenge} release a new large-scale multi-aspect multi-sentiment dataset and use capsule network building a strong baseline. Unlike these methods, we pay attention to the end-to-end ABSA task rather than the individual subtask, and propose a dynamic-length to dynamic-length routing algorithm, which can efficiently perform the multi-knowledge transfer.

\section{Conclusion}
In this paper, we propose an iterative multi-knowledge transfer network for the ABSA task, which can fully exploit the inter-task correlations among the three aspect-level subtasks with the proposed routing algorithm. Moreover, we design a more fine-grained method enabling our model to incorporate the document-level knowledge for pertinently enhancing the corresponding aspect-level tasks. Experimental results on three benchmark datasets demonstrate the effectiveness of our proposed approach, which yields state-of-the-art performance on most metrics.

\section*{Acknowledgements}
The research work descried in this paper has been supported by the National Key R\&D Program of China (2019YFB1405200) and the National Nature Science Foundation of China (No. 61976015, 61976016, 61876198 and 61370130). The authors would like to thank the anonymous reviewers for their valuable comments and suggestions to improve this paper.

\bibliography{anthology,custom}
\bibliographystyle{acl_natbib}

\appendix
\label{sec:appendix}
\section*{Appendix}
\section{Implementation Details}
Following~\cite{chen-qian-2020-relation}, we use 300d GloVe released by~\citeauthor{glove:14}~\shortcite{glove:14} as general-specific embeddings and the embeddings released by~\citeauthor{xu_acl2018}~\shortcite{xu_acl2018} as domain-specific embeddings. Our models are trained by Adam optimizer~\cite{Adam:14}, with learning rate $\eta_0 = 10^{-4}$, and batch size is set to 32. When training, we randomly sample 20\% of each training data as the validation set and the remaining 80\% as training set. 

We following~\cite{chen-qian-2020-relation} fix the domain-specific and general-specific word embeddings in all models, where the domain-specific embedding vectors are 100 dimensions. The trainable weight matrices in the {\em CNN} are initialized by following the Glorot Uniform strategy~\cite{pmlr-v9-glorot10a}. Besides, all biases are initialized as zero. We tune the number of {\em CNN} layer on the validation set of each dataset. Finally, The {\em CNN} layer number in the shared encoder is set to 2, and is fixed as 2, 2, 1 for the ATE subtask, the OTE subtask, and the ASC subtask in task-specific layers, respectively. The {\em CNN} layer in the shared encoder has 150 filters with kernel size k = 3 and 150 filters with kernel size k = 5. The {\em CNN} layers in each task-specific encoder have 300 filters with kernel size k = 5 per layer. The activation function is ReLu for each {\em CNN} layer. Dropout is employed after the embedding layer and each {\em CNN} layer, which is empirically set to 0.5. The discount coefficients $\lambda_1$, $\lambda_2$, $\lambda_3$, $\lambda_4$ and $\lambda_5$ in loss functions are not fine-tuned and empirically set to 1.0. 

Since the extracted aspect term may consist of several tokens and the predicted polarity of each token may be inconsistent, we thus following~\cite{chen-qian-2020-relation} only take the sentiment polarity of the first token of the current aspect term as the sentiment label for measuring the performance. We also note that only aspect terms have sentiment annotations and thus following~\cite{chen-qian-2020-relation} only consider ASC predictions on these aspect term-related tokens for computing the ASC loss and ignore the sentiments predicted on other tokens. 

For training, we first train the model with document-level tasks for five epochs, and then alternately train our model on aspect-level tasks with 2 epochs and document-level tasks with 1 epoch. Finally, we train the model for a fixed number of epochs, and obtain the best results at the epoch with the best F1-absa score on the validation set for producing the testing results, as did in~\cite{he_acl2019}. 

In our experiments, following~\cite{chen-qian-2020-relation}, we also use $BERT_{LARGE}$~\cite{bert} as the backbone to further investigate our model performance.

The neural model is implemented in Keras and all computations are done on an NVIDIA Tesla V100 GPU, where each experiment runs about 1$\sim$3 hours. 
Hyperparameter configurations for best-performing models have explained above. The method of choosing hyperparameter values is manual tuning on the validation and the criterion used to select is F1-absa. The downloadable version of used data can be found in: \url{https://github.com/ruidan/IMN-E2E-ABSA}, provided by IMN~\cite{he_acl2019}, where we use this data without any pre-processing.

\section{Experiments of Hyperparameters}
\label{H_P}
\textbf{Impact of Iteration Number: $T$.}

\noindent As an important hyperparameter, we investigate the impact of iterations $T$. \autoref{vary T} shows the change of F1-absa on the validation set of each dataset. We find that the best results can be obtained when $T$ equals 1, 2, and 4, respectively. There is no consistent conclusion about how to set this parameter. In general, $T$ is set to 1, 2, and 4 on D1, D2, and D3 in our experiments, respectively.

\begin{table}[h]
\centering
\small
\begin{tabular}{c|cccccc}
\toprule 
$T$&0&1&2&3&4&5\\\hline
D1 &62.78 &\bf{63.56} &63.14 &63.44 &63.00 &62.34\\
D2 &53.34 &55.25 &\bf{56.22} &{56.07} &55.47 &54.88\\
D3 &65.04 &{65.72} &{65.88} &65.72 &\bf{66.35} &65.78\\
\bottomrule
\end{tabular}
\caption{F1-absa (\%) scores with different $T$ values. Average results over 5 runs on the validation set are reported. }\label{vary T}
\end{table}

\medskip

\noindent\textbf{Impact of Routing Number: $iter$.}

\noindent \autoref{GCN_layers} (in the next page) shows the impact of the maximum number of the routing number $iter$ of the routing algorithm on the validation set of each dataset. The results demonstrate that the model achieves the best results when routing number equals 3 and further iterations do not further improve the performance. In general, the routing number is fixed to 3 in our experiments.
\begin{table}[h]
\centering
\small
\begin{tabular}{c|cccccc}
\toprule 
$iter$&1&2&3&4&5\\\hline
D1 &63.06 &63.80 &\bf{64.52} &64.02 &64.25 \\
D2 &56.28 &{56.47} &\bf{57.14} &56.70 &56.47 \\
D3 &65.71 &{66.32} &\bf{66.75} &{66.03} &66.00\\
\bottomrule
\end{tabular}
\caption{F1-absa (\%) scores with different routing number in Routing Block. Average results over 5 runs on the validation set are reported.}\label{GCN_layers}
\end{table}

\begin{table}[h]
\begin{center}
\scalebox{0.8}{
\setlength{\tabcolsep}{1.4mm}{
\begin{tabular}{c|c}
\toprule
{Sentence}  &\multicolumn{1}{l} {\makecell{The service is slow. }}  \\ \hline
{Aspect}    &  service \qquad \\ \hline
{Opinion}     &  slow \qquad  \\ \hline
{Sentiment Polarity} &  negative \qquad  \\\hline
{Aspect-Sentiment Pair} &  service-negative \qquad  \\\hline
{Aspect-Opinion Pair} &  service-slow \qquad  \\\hline
{Aspect-Opinion-Sentiment Triplet}& service-slow-negative \qquad  \\
\bottomrule
\end{tabular}}}
\end{center}
\caption{The example of Aspect-Sentiment Pair, Aspect-Opinion Pair, and Aspect-Opinion-Sentiment Triplet.}
\label{tbl:testE}
\end{table}

\section{Error Analysis}
We have checked some error examples and made a thorough error analysis, which can be roughly divided into 3 types. 1) Due to aspect extraction and opinion extraction are not always correctly identified, the Aspect-Opinion-Sentiment triplet is hard to handle. 2) The imbalanced label distribution in the training corpus. 3) The complex instances are hard to correctly deal with, such as the sentence that has multiple aspects and opinions, which are hardly effectively learned. For instance, in the sentence “coffee is a better deal than overpriced cosi sandwiches”, where two opinion terms “better” and “overpriced”, and two aspect terms “coffee” and “cosi sandwiches” are mentioned, where the sentiment polarities of them are “positive” and “negative”, respectively. In this case, our \textsc{IMKTN} correctly extracted all aspect terms, and the \textsc{IMKTN} successfully detected the opinion term “better” but failed to identify the opinion term “overpriced”, \emph{i.e.}, the OTE subtask failed partly, where the \textsc{IMKTN} made right sentiment classification for the aspect term “coffee” but assigned wrong sentiment polarity (“positive”) to the aspect term “cosi sandwiches”. The reason may be that the knowledge from the opinion term “better” contributed to the right sentiment classification for “coffee” but led to the wrong sentiment classification for “cosi sandwiches”. If the opinion term “overpriced” can be successfully identified, it may contribute to the right classification for “cosi sandwiches” with our routing algorithm. 

\end{document}